\documentclass[letterpaper, 10 pt, journal, twoside]{IEEEtran}

\usepackage{cite}
\usepackage{amsmath,amssymb,amsfonts}
\usepackage{graphicx}

\usepackage{float}
\usepackage{bm}
\usepackage{romannum}

\usepackage{booktabs}

\usepackage{algorithm}
\usepackage[noend]{algpseudocode}
\usepackage{textcomp}

\usepackage[flushleft]{threeparttable}

\newcommand{\etal}{\textit{et~al}. }
\usepackage[font=small,labelfont=bf]{caption}

\usepackage{hyperref}
\usepackage[dvipsnames]{xcolor}
\newcommand\myshade{85}
\colorlet{mylinkcolor}{violet}
\colorlet{mycitecolor}{YellowOrange}
\colorlet{myurlcolor}{Aquamarine}
\hypersetup{
  linkcolor  = mylinkcolor!\myshade!black,
  citecolor  = mycitecolor!\myshade!black,
  urlcolor   = myurlcolor!\myshade!black,
  colorlinks = true,
}

\begin{document}
\title{\LARGE \bf Learning Robotic Manipulation Tasks via Task Progress based Gaussian Reward and Loss Adjusted Exploration}

\author{Sulabh Kumra, Shirin Joshi and Ferat Sahin
\thanks{Manuscript received: August 12, 2021; Revised: October 24, 2021; Accepted: November 16, 2021.}
\thanks{This paper was recommended for publication by Editor J. Kober upon evaluation of the Associate Editor and Reviewers' comments.} 
\thanks{Sulabh Kumra is with OSARO Inc, San Francisco, CA 94103 USA and also with Rochester Institute of Technology, Rochester, NY 14623 USA.}
\thanks{Shirin Joshi is with Siemens Corporation, Corporate Technology, Berkeley, CA 94703 USA.}
\thanks{Ferat Sahin is with Multi-Agent Bio-Robotics Laboratory, Rochester Institute of Technology, Rochester, NY 14623 USA.}
\thanks{Corresponding author e-mail: \href{mailto:sk2881@rit.edu}{sk2881@rit.edu}}
\thanks{Digital Object Identifier (DOI): see top of this page.}
}


\markboth{IEEE Robotics and Automation Letters. Preprint Version. Accepted November 2021}
{Kumra \MakeLowercase{\textit{et al.}}: Learning Robotic Manipulation Tasks}

\maketitle

\begin{abstract}
Multi-step manipulation tasks in unstructured environments are extremely challenging for a robot to learn. Such tasks interlace high-level reasoning that consists of the expected states that can be attained to achieve an overall task and low-level reasoning that decides what actions will yield these states. We propose a model-free deep reinforcement learning method to learn multi-step manipulation tasks. We introduce a Robotic Manipulation Network (RoManNet), which is a vision-based model architecture, to learn the action-value functions and predict manipulation action candidates. We define a Task Progress based Gaussian (TPG) reward function that computes the reward based on actions that lead to successful motion primitives and progress towards the overall task goal. To balance the ratio of exploration/exploitation, we introduce a Loss Adjusted Exploration (LAE) policy that determines actions from the action candidates according to the Boltzmann distribution of loss estimates. We demonstrate the effectiveness of our approach by training RoManNet to learn several challenging multi-step robotic manipulation tasks in both simulation and real-world. Experimental results show that our method outperforms the existing methods and achieves state-of-the-art performance in terms of success rate and action efficiency. The ablation studies show that TPG and LAE are especially beneficial for tasks like multiple block stacking. Code is available at: \href{https://github.com/skumra/romannet}{https://github.com/skumra/romannet}
\end{abstract}



\section{Introduction}
\IEEEPARstart{R}{obotic} manipulation tasks have been the backbone of most industrial robotic applications, e.g. bin picking, assembly, palletizing, or machine tending operations. In structured scenarios, these tasks have been reliably performed by the methods used in the existing work \cite{bohg2013data, kopicki2016one}. While, in unstructured scenarios, simple tasks such as pick only tasks have been successfully performed using grasping approaches such as \cite{schmidt2018grasping, kumra2017robotic, yen2020learning}, complex tasks that involve multiple steps such as clearing a bin of mixed items, and creating a stack of multiple objects, remain a challenge.

Training end-to-end manipulation policies that map directly from image pixels to joint velocities can be computationally expensive and time exhaustive due to a large volume of sample space and can be difficult to adapt on physical setups \cite{levine2018learning, kalashnikov2018scalable, joshi2020robotic, florence2019self, Rajeswaran-RSS-18}. To solve this, many have tried pixel-wise parameterization of both state and action spaces, which enables the use of a neural network as a Q-function approximator \cite{zeng2018learning, hundt2020good, berscheid2019robot}. However, these approaches have a low success rate, long learning time, and cannot handle intricate tasks consisting of multiple steps and long horizons.


\begin{figure}
    \centering
    \includegraphics[width=0.96\linewidth]{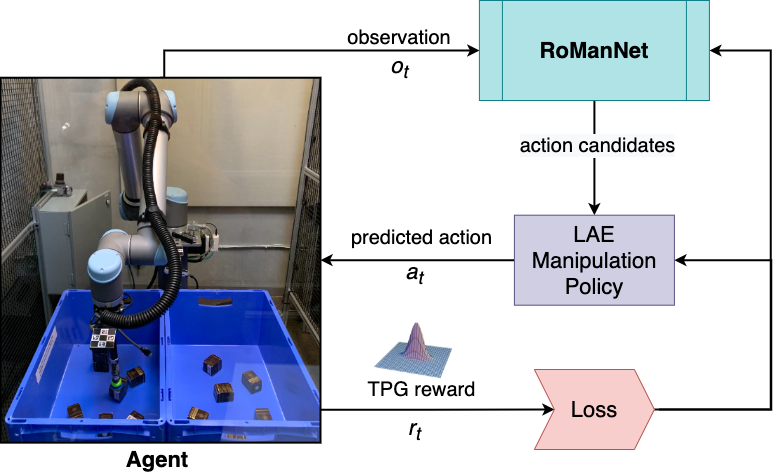}
    \caption{Proposed approach for training a vision-based deep reinforcement learning agent for efficiently learning multi-step manipulation tasks.}
    \label{fig: overview}
\end{figure}

In this work, we present a model-free deep reinforcement learning approach (shown in Fig. \ref{fig: overview}) to produce a deterministic policy that allows complex robot manipulation tasks to be effectively learned from pixel input. The policy directs a low-level controller to perform motion primitives rather than regressing motor torque vectors directly by learning a pixel-wise action success likelihood map. We propose a Task Progress based Gaussian (TPG) reward function to learn the coordinated behavior between intermediate actions and their consequences towards the advancement of an overall task goal. We introduce a Loss Adjusted Exploration (LAE) policy to explore the action space and exploit the knowledge, which helps in reducing the learning time and improving the action efficiency. The key contributions of this work are:
\begin{itemize}
    \item We introduce Robotic Manipulation Network (RoManNet), an end-to-end model architecture to efficiently learn the action-value functions and generate accurate action candidates from visual observation of the scene.
    \item We propose a TPG reward function that uses a sub-task indicator function and an overall task progress function to compute the reward for each action in a multi-step manipulation task.
    \item We address the challenge of balancing the ratio of exploration/exploitation by introducing an LAE manipulation policy that selects actions according to the Boltzmann distribution of loss estimates.
\end{itemize}
We demonstrate the effectiveness of our approach in simulation as well as in the real-world setting by training an agent to learn three vision-based multi-step robotic manipulation tasks. RoManNet trained with TPG reward and LAE policy performed significantly better than previous methods with only 2000 iterations in the real-world setting. We observed a pick success rate of 92\% for a mixed item bin-picking task and 84\% action efficiency for a block-stacking task.



\section{Related Work}
Robotic manipulation has always been an essential part of research in the field of robotics. There has been significant progress in recent years in robotic grasping that leverages deep learning and computer vision for generating grasp candidates for specific tasks that can be used to select suitable grasp poses for novel objects in fairly structured environments \cite{lenz2013deep,pinto2016supersizing,kumra2020antipodal}. While most prior manipulation methods focus on singular tasks, our work focuses on learning multi-step manipulation tasks which generalize to a wide range of objects and tasks.


Deep reinforcement learning can be used to learn complex robotic manipulation tasks by using model-free deep policies. Kalashnikov \etal demonstrate this by proposing a QT-opt technique to provide a scalable approach for vision-based robotic manipulation applications \cite{kalashnikov2018scalable}. Riedmiller \etal introduced a SAC-X method that learns complex tasks from scratch with the help of multiple sparse rewards where only the end goal is specified \cite{riedmiller2018learning}. The agent learned these tasks by exploring its observation space and the results showed that this technique was highly reliable and robust. More recently, Xu \etal introduced a learning-to-plan method called Deep Affordance Foresight (DAF), which learns partial environment models of affordances of
parameterized motor skills through trial-and-error \cite{xu2021deep}. Nematollahi \etal demonstrated that scene dynamics in the real-world can be learned for visuomotor control and planning \cite{nematollahi2020hindsight}.

Learning multi-step tasks with sparse rewards is particularly challenging because a solution is improbable through random exploration. Tasks such as clearing a bin with multiple objects \cite{mahler2017learning, kalashnikov2018scalable} do not include long-horizon and the likelihood of reverse progress is out of consideration. Many propose to use a model-free method through self-supervised learning. One such method proposed by Zeng \etal uses a visual pushing grasping (VPG) framework that can discover and learn to push and grasp through model-free deep Q learning \cite{zeng2018learning}. Similarly, Jeong \etal performed a stacking task by placing a cube over another cube using a two-stage self-supervised domain adaptation (SSDA) technique \cite{jeong2020self}. Zhu \etal presented a framework in which manipulation tasks were learned by using a deep visuomotor policy (DVP) that uses a combination of reinforcement learning and imitation learning to map RGB camera inputs directly into joint velocities \cite{zhu2018reinforcement}. Hundt \etal developed the SPOT framework \cite{hundt2020good}, that explores actions within the safety zones and can identify unsafe regions even without exploring and can prioritize its experience to only learn what is useful. Zeng \etal proposed Transporter Network trained using learning from demonstrations, which rearranges deep features to infer spatial displacements from visual input \cite{zeng2020transporter}. Kase \etal proposed a Deep Planning Domain Learning (DPDL) framework which learns a high-level model using sensor data to predict values for a large set of logical predicates consisting of the current symbolic world state and separately learns a low-level policy which translates symbolic operators into robust executable actions on a robot \cite{kase2020transferable}. DPDL framework worked well on manipulation tasks in a photorealistic kitchen scenario. Similarly, Driess \etal proposed a network architecture consisting of a high-level reasoning network, adaptation network, and low-level controllers to learn geometrically precise manipulation tasks for a dual robot arm system where the parameters of early actions are tightly coupled with those of later actions \cite{driess2021learning}. Kit assembly \cite{zakka2020form2fit, devgon2021kit}, and cloth manipulation \cite{borras2020grasping, seita2019deep} are some other tasks that involve multiple steps.

In our work, we focus on multi-step tasks that involve long-horizon planning and considers progress reversal. More closely related to our work is the VPG framework introduced by Zeng \etal which utilized a Fully Convolutional Network (FCN) as a function approximator to estimate the action-value function. However, this method had a low success rate and was sample inefficient. We introduce a sample efficient RoManNet framework to learn multi-step manipulation tasks, which is trained using a TPG reward function and LAE policy to address these problems. In addition to the grasp and pre-grasp primitives explored by Zeng \etal, we explore the placement primitive to evaluate our approach on multi-step manipulation tasks like building a stack of blocks.

\begin{figure*}
    \centering
    \vspace*{0.2cm}
    \includegraphics[width=0.94\textwidth]{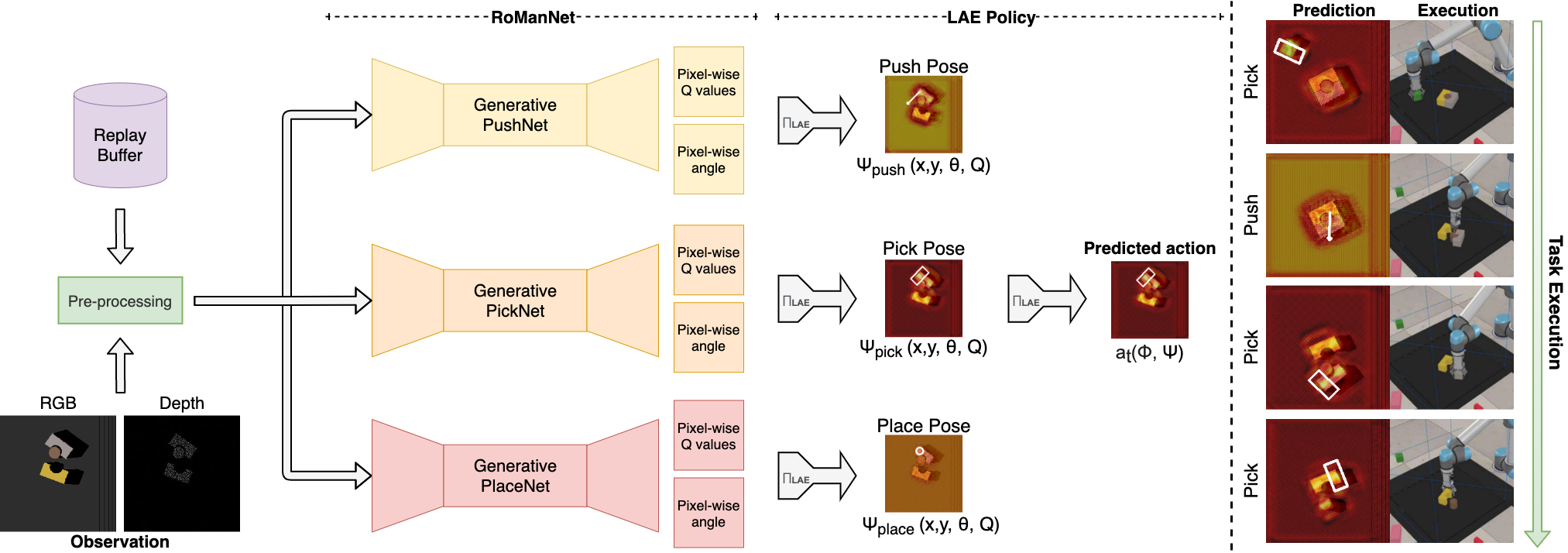}
    \caption{\textbf{Left}: Proposed Robotic Manipulation Network (RoManNet) based framework for learning multi-step manipulation tasks. The pre-processed (cropped, resized, and
normalized) inputs are fed into three generative networks which generate the action candidates. Each of these generative networks is a variant of GR-ConvNet. The LAE policy $\Pi_{LAE}$ selects an action that maximizes the expected reward. \textbf{Right}: An illustration of an agent using the learned robot manipulation policy to execute a multi-step manipulation task, which requires the robot to clear a bin with challenging arrangement using pushing and picking actions. The robot can be seen pushing the item to de-clutter the tight arrangement before picking.}
    \label{fig: romannet}
\end{figure*}

\section{Approach}
\label{sec: approach}


We consider the problem of efficiently learning multi-step robotic manipulation for unknown objects in an environment with unknown dynamics. Each manipulation task can be formulated as a Markov decision process where at any given state $s_t \in \mathcal{S}$ at time $t$, the robot makes an observation $o_t \in \mathcal{O}$ of the environment and executes an action $a_t \in \mathcal{A}$ based on policy $\pi(s_t)$ and receives an immediate reward of $\mathcal{R}_{a_t}( {s_t}, {s_{t+1}})$. In our formulation, $o_t \in \mathbb{R}^{4\times h \times w}$ is the visual observation of the robot workspace from RGB-D camera, and we divide the action space $\mathcal{A}$ into two components: action type $\Phi$ and action pose $\Psi$. The underlying assumption is that the edges of $o_t$ are the boundaries of the agent’s workspace, and $o_t$ embeds all necessary state information, thus providing sufficient information of the environment to choose the correct action.


We introduce RoManNet to approximate the action-value function $Q_\mu$, which predict manipulation action candidates $\mathcal{A}_{s_t}$ from the observation $o_t$ of the state $s_t$ at time $t$. The proposed LAE manipulation policy $\Pi_{LAE} (s_t, \mathcal{L}_t)$ determines the action $a_t$ from the action candidates $\mathcal{A}_{s_t}$ that maximizes the reward $\mathcal{R}$. Once the agent executes $a_t$, the reward is computed using a TPG reward function. The parameters $\mu$ are updated by minimizing the loss function. An overview of the proposed learning framework is illustrated in Fig. \ref{fig: romannet}.

\subsection{Manipulation Action Space}
We parameterize each manipulation action $a_t$ as two components: action type $\Phi$, which consists of three high-level motion primitives $\{push, pick, place\}$, and action pose $\Psi$, which is defined by the pose at which the action is performed. Each manipulation action pose in the robot frame is defined as:
\begin{equation}\label{eq: pose_robot}
    \Psi_r = (P, \theta_r, Q) 
\end{equation}
where, $P = (x, y, z)$ is the gripper’s center position, $\theta_r$ is the rotation of the gripper around the z-axis, and $Q$ is an affordance score that depicts the  “quality” of action. The manipulation action pose in image frame $\Psi_i$ is parameterized pixel-wise and defined as:
\begin{equation}
    \Psi_i = (x, y, \theta_i, Q)
\end{equation}
where $(x, y)$ is the center of action pose in image coordinates, $\theta_i$ is the rotation in the image frame, and $Q$ is the same affordance score as in equation \eqref{eq: pose_robot}.    

The high-level motion primitive behaviors $\Phi$ are defined as follows:
\begin{itemize}
    \item \textbf{Pushing:} $\Psi_{push} = (x, y, \theta, Q)$ denotes the starting pose of a 10cm push. A push starts with the gripper closed at $(x, y)$ and moves horizontally at a fixed distance along angle $\theta$.
    \item \textbf{Picking:} $\Psi_{{pick}} = (x, y, \theta, Q)$ denotes the middle position of a top-down grasp. During a pick attempt, both fingers attempt to move 3cm below $\Psi_{{pick}}$ (in the $-z$ direction) before closing the fingers.
    \item \textbf{Placing:} $\Psi_{place} = (x, y, \theta, Q)$ denotes the middle position of a top-down placement. During a place attempt, both fingers open when the place pose $\Psi_{place}$ is reached.
\end{itemize}

\begin{algorithm}
\caption{Learn action-value function $Q_\mu: \mathcal{S} \times \mathcal{A} \rightarrow \mathbb{R}$}
\label{alg:q-learning}
    \begin{algorithmic}
    \Require{Manipulation MDP with states $\mathcal{S}$ and actions $\mathcal{A}$, Transition function $\mathcal{T}: \mathcal{S} \times \mathcal{A} \rightarrow \mathcal{S}$, and Reward function $\mathcal{R}_{tpg}: \mathcal{S} \times \mathcal{A} \rightarrow \mathbb{R}$}
    \Procedure{Learning Policy}{$\mathcal{S}$, $\mathcal{A}$, $\mathcal{R}$, $\mathcal{X}$, $\mathcal{P}$}
        \State Initialize $Q_\mu$ with random weights
        \State Initialize experience replay buffer $\mathcal{D}$
        \While{$Q_\mu$ is not converged}
            \State Initialize sub-task indicator function $\mathcal{X}$
            \State Initialize overall  task progress function $\mathcal{P}$
            \State Start in state $s_t \in \mathcal{S}$
            \While{$s$ is not terminal}
                \State Receive the observation $o_t$ of the state $s_t$
                \State Calculate $\Pi_{LAE} (s_t, \mathcal{L}_t)$ according to eq. \ref{eq: policy}
                \State $a_t (\Phi, \Psi) \gets \Pi_{LAE} (s_t, \mathcal{L}_t)$
                \State Execute action on robot
                \State Obtain $\mathcal{X}$ and $\mathcal{P}$
                \State $r_t \gets \mathcal{R}_{tpg}(s_{t+1}, a_t)$ according to eq. \ref{eq: reward}
                \State Store transition $(s_t, a_t, r_t, s_{t+1})$ in $\mathcal{D}$
                \State $s_t \gets s_{t+1}$ according to $\mathcal{T}$
                \State Sample mini-batch from $\mathcal{D}$
                \State $Y_t \gets \mathbb{E}_\pi(s_{t+1}, a_t)$ according to eq. \ref{eq: expected reward}
                \State Update $Q_\mu$ minimising the loss $\mathcal{L}_t$ in eq. \ref{eq: loss}
            \EndWhile
        \EndWhile
        \Return $Q_\mu$
    \EndProcedure
    \end{algorithmic}
\end{algorithm}

\subsection{Learning the Action-Value Functions}
The proposed algorithm to learn the action-value function is described in algorithm \ref{alg:q-learning}. We approximate the action-value function using a neural network architecture we call RoManNet. In contrast to the DenseNet-121 fully connected networks as in \cite{zeng2018learning, hundt2020good}, it is built using three lightweight Generative Residual Convolutional Neural Network (GR-ConvNet) \cite{kumra2020antipodal} models PushNet, PickNet, and PlaceNet, one for each motion primitive behavior (pushing, picking, and placing respectively). Each individual GR-ConvNet model takes as input the image representation $o_t \in \mathbb{R}^{4 \times h \times w}$ of the state $s_t$ and generates a pixel-wise map of Q values $\mathcal{Q}_\Phi(s_t, a_t) \in \mathbb{R}^{h \times w}$ and corresponding rotation angle for each pixel $\theta_\Phi(s_t, a_t) \in \mathbb{R}^{h \times w}$ with the same image size and resolution as $o_t$. In this work, $h$ and $w$ are same and $\mathbb{R}^{h \times w}$ is represented as an image of resolution $224 \times 224$. The overall Q-function $Q_{\mu} (s_t, a_t)$ selects the action $a_t$ that maximizes the Q-value over pixel-wise maps for all manipulation action poses: $\mathrm{argmax}\, Q_{\mu} (s_t, a_t) = \mathrm{argmax}\, (\mathcal{Q}_{push}(s_t), \mathcal{Q}_{pick}(s_t), \mathcal{Q}_{place}(s_t))$, which gives the center position $(x, y)$ of the action in image coordinates and corresponding pixel in the rotation angle pixel-wise map $\theta_\Phi(s_t, a_t)$ gives the rotation of the gripper $\theta_i$.


The RoManNet is continuously trained to approximate the optimal policy with prioritized experience replay using stochastic rank-based prioritization and leverages future knowledge via a recursively defined expected reward function:
\begin{equation}\label{eq: expected reward}
    \mathbb{E}_\pi(s_{t+1}, a_t) = 
    \mathcal{R}(s_{t+1}, a_t) + \eta (\gamma \mathcal{R}(s_{t+2}, a_{t+1}))
\end{equation}
\noindent where, $\gamma \in [0, 1)$ is the discount factor and $\eta$ is a reward propagation factor. A value of $\gamma$ closer to $0$ indicates that the agent will choose actions based only on the current reward and a value approaching $1$ indicates that the agent will choose actions based on the long-term reward. The reward propagation factor $\eta$ ensures that future rewards only propagate across time steps where subtasks are completed successfully. The value of $\eta$ at time step $t$ equals to $1$ if $\mathcal{R}(s_{t+1}, a_t) > 0$, and $0$ otherwise.

The loss of the network is computed using the Huber loss function at each time step $t$ as:
\begin{equation}\label{eq: loss}
    \mathcal{L}_t = \begin{cases}
    \frac{1}{2} \delta(t)^2 & \text{for }|\delta(t)| \le 1, \\
    \left(|\delta(t)|-\frac{1}{2}\right) & \text{otherwise.}
    \end{cases}
\end{equation}
where $\delta(t) = Q_{\mu_t}(s_t, a_t) - Y_t$ is the TD-error and $\mu_t$ are the weights of RoManNet at time step $t$. We pass the gradients only through the network from which the predictions of the executed action $a_t$ were computed. For an antipodal gripper, the pick and place motion primitives are symmetrical around 0-180\textdegree\ , thus we can pass the gradients for the opposite orientation as well. The models are trained using Stochastic Gradient Descent with a fixed learning rate of $10^{-4}$, a momentum of $0.9$, and weight decay of $2^{-5}$. The models are continuously trained on past trials using prioritized experience replay with future discount $\gamma = 0.5$.

\subsection{Task Progress based Gaussian Reward}
The reward function $\mathcal{R}(s_{t+1}, a_t) \in \mathbb{R}^{h \times w}$ operates on two principles: actions which advance overall task progress receive a reward proportional to the quantity of progress, but actions which reverse the progress receive 0 reward. The task progress is measured using: (\romannum{1}) a sub-task indicator function $\mathcal{X}(s_{t+1}, a_t)$, which equals to $1$ if $a_t$ leads to a successful primitive action and $0$ otherwise, and (\romannum{2}) an overall task progress function $\mathcal{P}(s_{t+1}, a_t) \in [0, 1]$, which is proportional to the progress towards an overall goal. We define our task progress based reward function as:
\begin{equation}
    \mathcal{R}_{tp}(s_{t+1}, a_t) = \mathcal{W}(\Phi) \mathcal{X}(s_{t+1}, a_t) \mathcal{P}(s_{t+1}, a_t)
\end{equation}
\noindent where $\mathcal{W}(\Phi)$ is a weighting function that depends on the primitive motion action type $\Phi$. In this work, we set $\mathcal{W}(\Phi)$ to 0.2 for pushing action, and 1 for picking and placing actions. In our experiments, to compute the sub-task indicator $\mathcal{X}(s_{t+1}, a_t)$, a push action is successful if it perturbs an object, a pick action is successful if an object is grasped and raised from the surface, and a place action is successful only if it increases the stack height. The overall task progress function $\mathcal{P}(s_{t+1}, a_t)$ is computed using the depth image. For item removal related tasks, it is the ratio of number of occupied pixels to the total number of pixels, and for the stacking task it is proportional to the height of the stack.

The task progress based reward function $\mathcal{R}_{tp}$ is a single pixel in $\mathbb{R}^{h \times w}$ and to make the reward function more robust, we smooth it using an anisotropic Gaussian distribution \cite{tsiotsios2013choice} parameterized with standard deviations $\sigma_x$ and $\sigma_y$. We used an anisotropy ratio of 2 (i.e., $\sigma_x / \sigma_y = 2$), where $x$ and $y$ are the axis of the anisotropic Gaussian distribution and $x$ is aligned with the $x$-axis of the gripper. The proposed Task Progress based Gaussian (TPG) reward function is specified as follows:
\begin{equation}
    G(x, y, \sigma_x, \sigma_y) = 
    \frac{1}{{2\pi \sigma_x \sigma_y}}e^{{ - \left( \frac{x^2}{2 \sigma_x ^2} + \frac{y^2}{2 \sigma_y ^2} \right) }}
\end{equation}
\begin{equation}
    \mathcal{R}_{g}(s_{t+1}, a_t) = \mathcal{R}_{tp}(s_{t+1}, a_t) \circledast G(x, y, \sigma_x, \sigma_y)
\end{equation}
\begin{equation}\label{eq: reward}
    \mathcal{R}_{tpg}(s_{t+1}, a_t) = \max ( \mathcal{R}_{tp}(s_{t+1}, a_t), \mathcal{R}_{g}(s_{t+1}, a_t))
\end{equation}
where $\circledast$ is the convolution operator, and $G$ is the anisotropic Gaussian filter applied to $\mathcal{R}_{tp}$. The intuition here is that action poses in the local neighbourhood should yield similar reward. We experimentally show the advantages of using $\mathcal{R}_{tpg}$ as compared to $\mathcal{R}_{tp}$ in section \ref{sec: ablation}.

\subsection{Loss Adjusted Exploration Policy}
We introduce a Loss Adjusted Exploration (LAE) policy to reduce unnecessary exploration once knowledge about initial states has been sufficiently established. The LAE policy extends the $\epsilon$-greedy policy similar to the Value-Difference based Exploration \cite{tokic2010adaptive}, and eliminates the need for hand-tuning the exploration rate $\epsilon$ as in \cite{zeng2018learning, hundt2020good}. We define the update steps for loss dependent exploration probability $\mathcal{E} (\mathcal{L}_t)$ as the following Boltzmann distribution of loss estimates:
\begin{equation}
    f(\mathcal{L}_t, \sigma) = \frac{1-e^(\frac{-|\alpha \cdot \mathcal{L}_t|}{\sigma})}
    {1+e^(\frac{-|\alpha \cdot \mathcal{L}_t|}{\sigma})}
\end{equation}

\begin{equation}
    \mathcal{E}_{t+1} (\mathcal{L}_t) = \beta \cdot f(\mathcal{L}_t, \sigma) + (1-\beta) \cdot \mathcal{E}_t (\mathcal{L}_t)
\end{equation}

\noindent where $\mathcal{L}_t$ is the loss of the network computed using equation \eqref{eq: loss} at each time step $t$, $\sigma$ is a positive constant called inverse sensitivity, $\alpha$ is a constant set to $0.5$ in our work, and $\beta \in [0,1)$ is a parameter determining the influence of the selected action on the exploration rate. The LAE policy is defined as:
\begin{equation}\label{eq: policy}
    \Pi_{LAE}({s_t}, \mathcal{L}_t) = \begin{cases}
    \mathcal{U} (\mathcal{A}_{s_t}) & \text{if } \xi < \mathcal{E} (\mathcal{L}_t), \\
    \underset{a_t \in \mathcal{A}(s_t)}{\mathrm{argmax}}\, Q_\mu (s_t, a_t)              & \text{otherwise.}
    \end{cases}
\end{equation}
\noindent where $\mathcal{U}$ is a uniform distribution over action candidates $\mathcal{A}_{s_t}$ and $\xi \in [0,1)$ is a random number drawn at each time step $t$ from a uniform distribution.

\begin{figure*}
    \centering
    \vspace*{0.2cm}
    \includegraphics[width=0.96\linewidth]{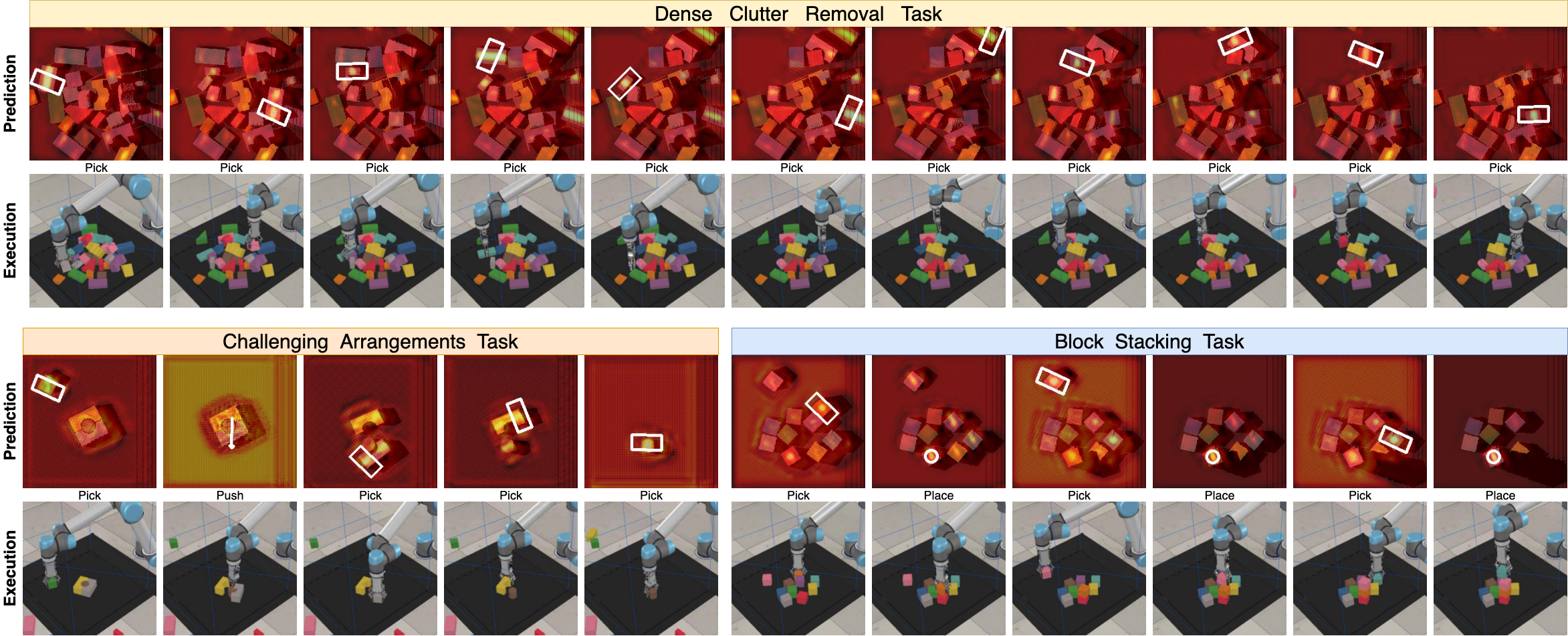}
    \caption{Visualization of various tasks being executed in simulation using our trained model. The first one is the dense clutter removal task, where the robot can be seen picking objects from the clutter of 30 objects in a bin and removing them from the bin one after another. We observe that the robot first picks any objects that are away from the clutter. The second is one of the challenging arrangements which has tightly packed objects. We observe that the robot first pushes the packed objects to de-clutter and then picks them up. The third is the block stacking task with a goal stack height of 4. We can see the robot performing consecutive pick and place actions to build a stack using 10 cubes randomly placed in the bin. The robot learns not to pick blocks from the stack being built as it leads to progress reversal and thus a lower reward.}
    \label{fig: qualitative_results}
\end{figure*}

\section{Experiments}

We conduct experiments in both simulated and real settings to evaluate the proposed method across various tasks. We design our experiments to investigate the following three questions: (\romannum{1}) How well does our method perform on different multi-step manipulation tasks? (\romannum{2}) Does our method improve task performance as compared to baseline methods? (\romannum{3}) What are the effects of the individual components of our framework in solving multi-step manipulation tasks, i.e. without the Loss Adjusted Exploration or Task Progress based Gaussian reward?


\subsection{Experimental Setups}
For the simulated environment, we expanded on the CoppeliaSim based setup used in \cite{zeng2018learning} to provide a consistent environment for fair comparisons and ablations. The environment simulates the agent using a UR5 robot arm with an RG2 gripper. Bullet Physics 2.83 is used to simulate the dynamics and CoppeliaSim’s internal inverse kinematics module for robot motion planning. A statically mounted perspective 3D camera is simulated in the environment to capture the observations of the states. Color and depth images of size $640 \times 480$ are rendered with OpenGL using the simulated camera. For the real-world setup, we used a UR10 robot with a Piab suction cup as the end effector. The RGB-D images of the scene are captured using an Intel RealSense D415 mounted rigidly above the workspace.

\subsection{Evaluation Metrics}
For evaluating the trained model, the policy is greedy deterministic and the model weights are reset to the trained weights at the start of each new test run. For each of the test cases, we execute 30 runs with new random seeds and evaluate performance with the following metrics found in \cite{zeng2018learning, hundt2020good}:
\begin{itemize}
    \item Completion rate: the average percentage runs in which the policy completed the given task without 10 consecutive fail attempts.
    \item Pick success rate: the average percentage of object picking success per completion.
    \item Action efficiency: a ratio of the ideal to the actual number of actions taken to complete the given task.
\end{itemize}

\begin{figure*}
    \centering
    \vspace*{0.2cm}
    \includegraphics[width=0.94\linewidth]{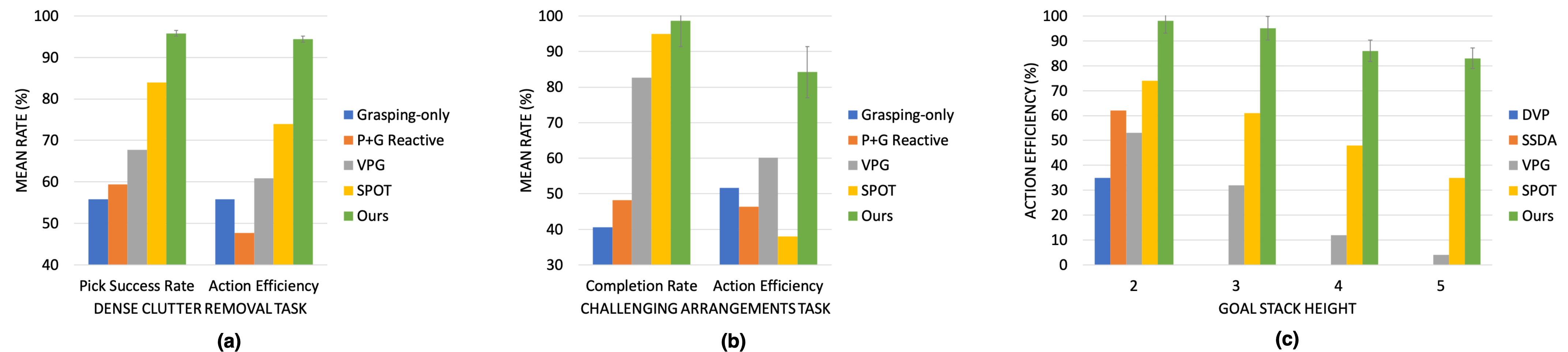}
    \caption{Quantitative comparisons with prior work for various tasks in similar simulation setups. Results for DVP and SSDA methods are borrowed from respective papers. All other results are reproduced in simulation using the open source code and models. (a) Performance for Dense Clutter Removal Task in terms of pick success rate and action efficiency. (b) Performance for Challenging Arrangements in terms of mean completion rate and action efficiency. (c) Performance for block stacking task in terms of mean action efficiency for various goal stack heights.}
    \label{fig: allresults}
\end{figure*}

\subsection{Simulation Experiments}
We design three manipulation tasks to evaluate our method in simulation: dense clutter removal, clearing 11 challenging test cases with adversarial objects, and stacking multiple objects. All tasks share the same MDP formulation in section \ref{sec: approach}, while the object set, manipulation action space, and the reward function are different for each task. Three sets of objects are used in these tasks: (\romannum{1}) 9 different 3D toy blocks (same as VPG \cite{zeng2018learning}), (\romannum{2}) 49 diverse evaluation objects from EGAD \cite{morrison2020egad}, and (\romannum{3}) cubes only for stacking task. To train our models for all tasks, objects with randomly selected shapes and colors are dropped in front of the robot at the start of each experiment. The robot then automatically performs data collection by trial and error. For object removal-related tasks in simulation, objects are removed from the scene after a successful pick action. The environment is reset at the termination of the task.

\textbf{Dense Clutter Removal Task:}
We first evaluate the proposed method in the simulated environment where 30 objects are randomly dropped in a bin. The goal of the agent is to remove all objects from the bin in front of the robot by executing pushing or picking actions. We trained the agent with only 10 objects instead of 30 objects. This helps to test the generalization of policies to more cluttered scenarios. The first experiment compares the proposed method to previous methods in simulation using the adversarial objects from \cite{zeng2018learning}. Comparison of our results with Grasping-only \cite{zeng2018robotic}, P+G Reactive \cite{zeng2018learning}, VPG \cite{zeng2018learning} and SPOT \cite{hundt2020good} are summarized in Fig. \ref{fig: allresults}(a). We see that with a pick success rate of 95.8\%, the proposed method outperforms all other methods across all metrics. We observe that our proposed method learned to pick items from a dense clutter without having to declutter them before picking. We hypothesize that this led to the high action efficiency of 94.4\%. The second experiment was with even more complex objects from the validation set of EGAD \cite{morrison2020egad}. We observed a high success rate of 92.2\% during testing on these complex objects, which suggests good generalizability of our proposed method.



\textbf{Challenging Arrangements Task:}
We also compare the proposed method with other methods in simulation on 11 challenging test cases with adversarial clutter from \cite{zeng2018learning}. These test cases are manually engineered to reflect challenging picking scenarios which allow us to evaluate the model’s robustness. Each test case consists of a unique configuration of 3 to 6 objects placed in a tightly packed arrangement that will be challenging even for an optimal picking-only policy as it will require de-cluttering them before picking. As a sanity check, a single isolated object is additionally placed separately from the main configuration. Similar to the dense clutter removal task, we trained a policy that learns to push and pick with 10 objects randomly placed in a bin and tested it on 11 challenging test cases. Performance comparison with previous work is shown in Fig. \ref{fig: allresults}(b). Across the collection of test cases, we observe that our proposed method can successfully solve 10/11 cases with a 100\% completion rate and an overall completion rate of 98.6\%. Moreover, the action efficiency of 84.2\% with our method indicates that our method is significantly better than VPG \cite{zeng2018learning} and SPOT \cite{hundt2020good} methods that attained action efficiencies of only 60.1\% and 38\% respectively.

\textbf{Block Stacking Task:}
To truly evaluate our multi-step task learning method, we consider the task of stacking multiple blocks on top of each other. This constitutes a challenging multi-step robotic task as it requires the agent to acquire several core abilities: picking a block from 10 blocks arbitrarily placed in the bin, precisely placing it on top of the second block, and repeating this process until a goal stack height is reached. We performed multiple experiments with a goal stack height in the range of 2 to 5. Fig. \ref{fig: allresults}(c) summarises the performance of our method compared to prior work. Jeong \etal used self-supervised domain adaptation (SSDA) \cite{jeong2020self} and Zhu \etal used deep visuomotor policy (DVP) \cite{zhu2018reinforcement} and tested it with a goal stack height of 2. Hundt \etal used SPOT \cite{hundt2020good} and tested it with a goal stack height of 4. For this task with the highest complexity, our approach seems to perform significantly better with the action efficiency of 98\% and 86\% for a goal stack height of 2 and 4, respectively. A possible reason is that, although the task is very complex, the TPG reward function helps learn optimal multi-step manipulation policy.


\begin{figure*}
    \centering
    \vspace*{0.2cm}
    \includegraphics[width=0.96\linewidth]{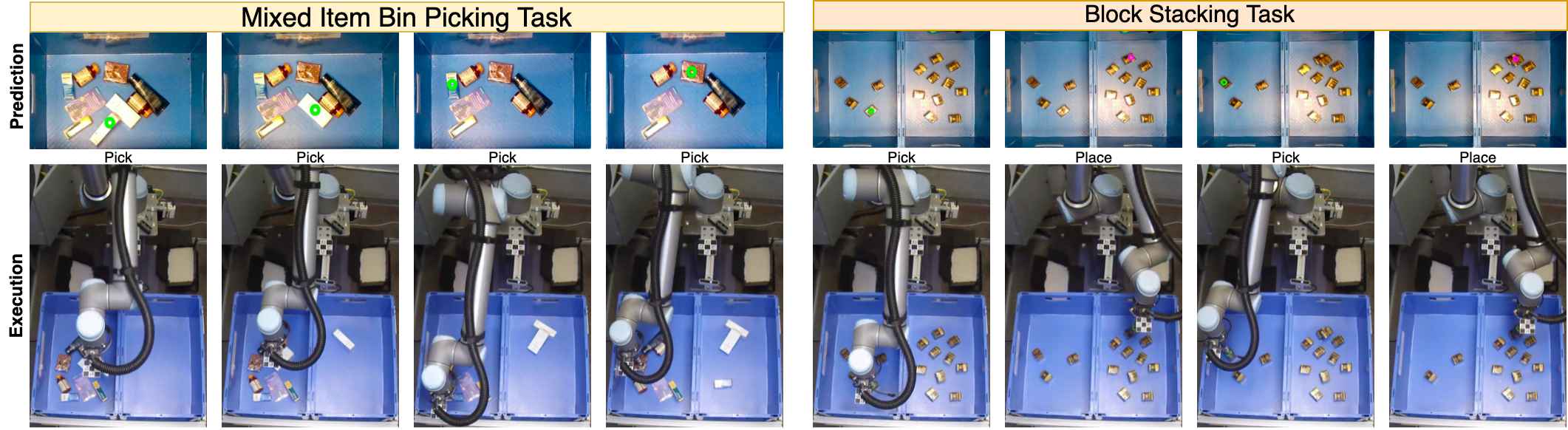}
    \caption{Illustration of mixed item bin picking and block stacking tasks being executed by the UR10 robot using the model trained for approximately 4 hours in the real-world. We provide additional qualitative results in the attached video.}
    \label{fig: ur-real}
\end{figure*}

\begin{table}
\begin{center}
\caption{Comparison of performance of state-of-the-art reinforcement learning based grasping methods in real-world settings}
\label{tab: real-grasping}
\begin{tabular}{l c c c}
\toprule
\textbf{Approach} & \textbf{Success \%} & \textbf{Training Steps} & \textbf{Test Items}\\
\midrule
VPG \cite{zeng2018learning} & 68 & 2.5k & 20 seen\\
SPOT \cite{hundt2020good} & 75 & 1k & 20 seen \\
Levine \etal \cite{levine2018learning} & 78 & 900k & - \\
QT-Opt \cite{kalashnikov2018scalable} & 88 & 580k & 28 seen \\
Grasp-Q-Network \cite{joshi2020robotic} & 89 & 7k & 9 seen \\
Berscheid \etal \cite{berscheid2019robot} & 92 & 27.5k & 20 seen\\
\midrule
RoManNet & \textbf{96 $\pm$ 1} & 2k & 20 seen \\
RoManNet & \textbf{92 $\pm$ 3} & 2k & 10 unseen \\
\bottomrule
\end{tabular}
\end{center}
\end{table}

\subsection{Real-World Experiments}
We validate our method in the real-world on two tasks: mixed item bin picking and block stacking. Fig. \ref{fig: ur-real} illustrates the two tasks being executed by the UR10 robot using RoManNet trained for 2k iterations ($\sim4$ hours of robot run time).

\textbf{Mixed Item Bin Picking:} For the mixed item bin-picking task, the robot picks items from the source bin and places them into a destination bin until the source bin is empty, and then swaps the source and destination bin for the next run. 30 different items were used for training the model. To automate the training process, suction feedback is used to detect a successful pick and a binary classifier is used to detect if the bin is empty after each manipulation action. We observe that the robot adopts diverse strategies to clear the clutter. The performance results in Table \ref{tab: real-grasping} shows that RoManNet performs consistently better than all previous methods. Pick success rate of 96\% after training for only 2k iterations demonstrates the high sample efficiency of our method. Generalization is a key index for applications in industrial and logistics automation. Our experiments with 10 unseen items show that our method is extremely impressive as it can generalize to novel objects in a real-world setting and still achieve a success rate of 92\%.


\textbf{Real-World Block Stacking:} The real-world block stacking task is similar to the one in simulation, where the robot performs manipulation actions until the goal stack height is reached. The depth measurements from the overhead camera are used to determine the current stack height. This task is particularly challenging as the agent needs to learn to align the position and orientation of the picked item with the stack during placement. Remarkably, we observed a 100\% completion rate and 84\% action efficiency for a goal stack height of 4, outperforming the state-of-the-art action efficiency of 61\% reported by Hundt \etal in \cite{hundt2020good}.

\begin{figure}
    \centering
    \includegraphics[width=0.98\linewidth]{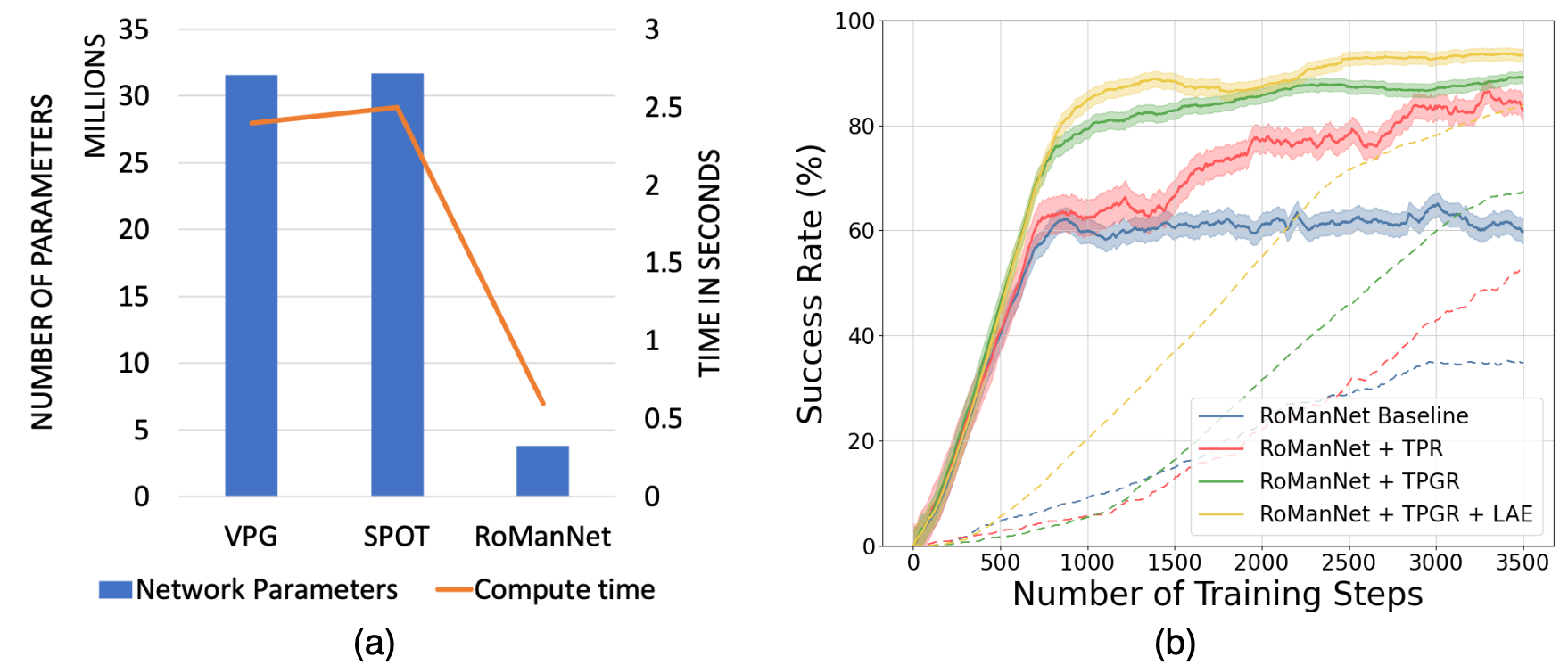}
    \caption{Ablation studies. (a) Comparison of network parameters and compute time of RoManNet with prior work (b) Learning curves for ablation of techniques used in conjunction with RoManNet for training agent on block stacking task. Solid lines indicate mean action success rates and dotted lines indicate mean action efficiency over training steps.}
    \label{fig: ablation}
\end{figure}

\subsection{Ablation Studies}
\label{sec: ablation}
In order to assess the necessity and efficacy of the different components of our proposed method, described in section \ref{sec: approach}, we provide ablation experimental results. Fig. \ref{fig: ablation}(a) shows a comparison of the number of network parameters and compute time for calculating the Q-values for a pushing and placing task of RoManNet with network architectures used in VPG \cite{zeng2018learning} and SPOT \cite{hundt2020good}. As RoManNet has significantly fewer network parameters, the compute time is significantly smaller (0.6 sec), which makes RoManNet more suitable for real-world manipulation tasks. Fig. \ref{fig: ablation}(b) compares the learning curves for the underlying algorithm against the baseline approaches for the complex multi-step task of block stacking.

\textbf{RoManNet Baseline}: We establish a baseline using the primary reward function found in VPG \cite{zeng2018learning} and $\epsilon$-greedy exploration policy. We define the baseline reward function as function of sub-task indicator only, i.e $\mathcal{R}_{base}(s_{t+1}, a_t) = \mathcal{X}(s_{t+1}, a_t)$. The only case where this baseline model is on par with the full model is the clutter removal task, in which both the baseline and the full model achieved similar levels of performance. We hypothesize that this is due to the short length of the task, where the task progress-based reward did not play a significant role.

\textbf{RoManNet + TPGR}: The second baseline is a combination of the TPG reward function and $\epsilon$-greedy exploration policy. We observe that TPG reward helped the agent learn to avoid task progress reversal for the multi-step block stacking task, thus improving the success rate by more than 25\%. We also observed a significant improvement in action efficiency for the challenging arrangements and block stacking tasks. Moreover, we observed that anisotropic gaussian smoothing improved manipulation action stability by removing maxima that were close to regions of low Q values. For example, RoManNet generated more stable action poses when trained with $\mathcal{R}_{tpg}$ as compared to $\mathcal{R}_{tp}$. As seen in Fig. \ref{fig: ablation}(b), training with $\mathcal{R}_{tpg}$ reward function improved the action success rate by 9\% compared to training with $\mathcal{R}_{tp}$ for block stacking task.

\textbf{RoManNet + TPGR + LAE}: For the complete proposed model, the proposed LAE policy is used instead of the $\epsilon$-greedy exploration. We observe an 11\% improvement in action efficiency for the block stacking task when RoManNet is trained with LAE policy instead of the baseline $\epsilon$-greedy policy. This suggests that LAE policy improves the efficiency of exploration, which is critical in real-world applications as training for a large number of iterations on a physical robot is expensive.

\section{Conclusions}
In this work, we present a vision-based deep reinforcement learning framework to effectively learn complex manipulation tasks that consist of multiple steps and long-horizon planning. The experimental results indicate that our TPG reward which computes reward based on the actions that lead to successful motion primitives and progress toward the overall task goal can successfully handle progress reversal in multi-step tasks. Moreover, we showed that our LAE policy can be used to curb unnecessary exploration which can occur after the initial states have already been explored. Compared to the previous work in multi-step manipulation, empirical results demonstrate that our method outperformed all previous methods in various multi-step tasks in simulation as well as real-world settings. We demonstrate a high generalization ability of our method by showing that our method achieves state-of-the-art results in performing multi-step tasks on previously unseen objects.





\bibliographystyle{IEEEtran} 
\bibliography{references} 


\end{document}